\documentclass[final,5p,times,twocolumn]{elsarticle}

\usepackage{subcaption}
\usepackage{balance}
\usepackage{framed,multirow}
\usepackage{array}

\newcolumntype{L}[1]{>{\raggedright\let\newline\\\arraybackslash\hspace{0pt}}m{#1}}
\newcolumntype{C}[1]{>{\centering\let\newline\\\arraybackslash\hspace{0pt}}m{#1}}
\newcolumntype{R}[1]{>{\raggedleft\let\newline\\\arraybackslash\hspace{0pt}}m{#1}}

\usepackage{hyperref}
\usepackage{tablefootnote}

\usepackage{adjustbox}

\makeatletter
\def\ps@pprintTitle{%
	\let\@oddhead\@empty
	\let\@evenhead\@empty
	\def\@oddfoot{\footnotesize\itshape Published in Computer Vision and Image Understanding\hfill \today}%
	\let\@evenfoot\@oddfoot}
\makeatother

\journal{Computer Vision and Image Understanding}

\begin{document}

\begin{frontmatter}

\title{Predicting the Future from First Person (Egocentric) Vision: A Survey}

\author[1]{Ivan Rodin\fnref{firstfoot}}
\author[1]{Antonino Furnari\fnref{firstfoot}}
\author[2]{Dimitrios Mavroedis}
\author[1]{Giovanni Maria Farinella\corref{cor1}\fnref{firstfoot} }
\cortext[cor1]{Corresponding author:}
\ead{gfarinella@dmi.unict.it}

\address[1]{Department of Mathematics and Informatics, University of Catania, V. Andrea Doria, Catania 95125, Italy}
\address[2]{Philips Research, High Tech Campus 34, 5656 Eindhoven, Netherlands}



\fntext[firstfoot]{These authors are co-first authors and contributed equally to this work.}

\begin{abstract}
Egocentric videos can bring a lot of information about how humans perceive the world and interact with the environment, which can be beneficial for the analysis of human behaviour. The research in egocentric video analysis is developing rapidly thanks to the increasing availability of wearable devices and the opportunities offered by new large-scale egocentric datasets. As computer vision techniques continue to develop at an increasing pace, the tasks related to the prediction of future are starting to evolve from the need of understanding the present. Predicting future human activities, trajectories and interactions with objects is crucial in applications such as human-robot interaction, assistive wearable technologies for both industrial and daily living scenarios, entertainment and virtual or augmented reality. This survey summarises the evolution of studies in the context of future prediction from egocentric vision making an overview of applications, devices, existing problems, commonly used datasets, models and input modalities. Our analysis highlights that methods for future prediction from egocentric vision can have a significant impact in a range of applications and that further research efforts should be devoted to the standardisation of tasks and the proposal of datasets considering real-world scenarios such as the ones with an industrial vocation.
\end{abstract}


\end{frontmatter}


\section{Introduction}
\label{sec:introduction}


Wearable cameras allow to collect visual information from the human perspective. The analysis of such data through egocentric (first-person) vision allows to study human behaviour in a much more direct manner. Indeed, in egocentric videos camera movements are usually based on the wearer's intentions and activities, and the manipulated objects are usually clearly visible in the frame and not hidden by other objects or people. This unique point of view makes it possible not only to recognize the activities performed by the camera wearer, but also to predict their future intentions and behaviour. Predicting the future from egocentric video is a relatively new task which has already found many applications in assistive technologies \citep{ohn2018personalized}, robotics \citep{soo2016egocentric}, entertainment \citep{liang2015ar, taylor2020towards} and autonomous vehicles \citep{hirakawa2018survey}.

{The psychology and computer vision communities have used an extensive vocabulary to describe tasks related to the modelling of the future. However, there is no common agreement on which terms to use to describe various aspects of predicting the future. As an example, it is common to find works using the term \textit{``action anticipation''} to refer to the tasks of predicting an on-going action \citep{sadegh2017encouraging, rodriguez2018action}, an action in the near future \citep{koppula2015anticipating, gao2017red, furnari2019would}, or an action in the distant future \citep{abu2018will, ke2019time}. We believe that a unified vocabulary is desirable to avoid misunderstandings and discrepancies, which arise naturally when new research problems are formulated. To this aim we propose to consider the terminology previously introduced by works in psychology, with the aim to provide a principled taxonomy, which may be beneficial for future works.}

{Specifically, we consider the terminology introduced by \citep{bubic2010prediction} to define the taxonomy introduced in Table~\ref{tab:definitions}. \citep{bubic2010prediction} use the term \textit{prediction} to describe both an event expected to happen in a certain context, and the overall process of making such predictions. Thus, we use the term ``future prediction'' to describe the general concept of predicting future events, which generalises a wide range of tasks.}

\begin{table}[t]\centering
\caption{Definitions for egocentric future prediction}\label{tab:definitions}
\scriptsize

\begin{tabular}{|L{1.4cm}|L{6.5cm}|}
\hline
\textbf{Term} & \textbf{Definition}\\ \hline

Prediction & The general concept of predicting future events, which generalises a wide range of future-related tasks.  \\ \hline

Anticipation & The process of formulating and communicating short-term expectations to sensory or motor areas. \\ \hline

Expectation & A representation of what is predicted to occur in the future (short-term). \\ \hline

Prospection & The process of predicting potential distant future events (long-term). \\ \hline

Forecasting & The tasks of modelling future events based on time-series data. \\ \hline

\end{tabular}

\end{table}

The term \textit{expectation} will be used to refer to a representation of what is predicted to occur in the future. The term \textit{anticipation} can be used to describe the process of formulating and communicating short-term expectations to sensory or motor areas. In other words, anticipation is a preparatory process of formulating hypotheses about the future. While anticipation and expectation are used in the context of short-term future predictions, the term \textit{prospection} is used for potential distant future events. One can imagine a person performing a natural activity, such as cooking a soup. The person already cut the vegetables and now they are going to add them to the saucepan with boiling water. In this setting, formulating the hypothesis that the person will add vegetables to the pan is \emph{anticipation}, an image of vegetables boiling in the water is an {expectation}, whereas predicting soup eating is prospection. The term \textit{forecasting} will be used to describe events based on time-series data. Note that, according to this definition, this temporal dimension forms the main difference between \textit{forecasting} and \textit{prediction}. We believe that the proposed taxonomy is principled and operationally effective because: 1) it is grounded in the psychology literature, 2) it disambiguates the process of formulating predictions (e.g., anticipation, prospection, forecasting) from the outcome of such process (e.g., expectation), 3) it binds definition to the temporal scale of the prediction (e.g., short vs long-term), which is crucial as discussed in the following.


Predictive methods may differ significantly from one another. In the following, we revise a series of parameters which can be used to classify general predictive systems as they play an important role in perception~\citep{pezzulo2008coordinating}:


\textit{Time-scale of Prediction:} The temporal horizon of the prediction influences its accuracy. While comparing and evaluating predictive systems, it is important to take into account the time-scale of predictions. A trade-off between the time-scale of prediction and the quality of predictions is generally required. Given the uncertainty of future predictions, a good model that aims to make long-term predictions should also 
report the inferred time-scale of prediction, in other words, answering to the question \textit{``When?''} in addition to the question \textit{``What?''};

\textit{Focusing Capabilities:} Focusing is important when processing egocentric videos. Due to the wearable nature of the camera, egocentric videos tend to include highly variable visual content. 
Hence, the video may contain elements or objects which are not relevant to the future prediction. Thus, a predictive system should be able to ignore irrelevant elements and focus only on the relevant parts of the surrounding environment;

\textit{Generalization Capabilities:} Predictive systems should generalise predictions over similar events or inputs. This is particularly relevant for future predictions, as methods should be able to formulate hypotheses about future events which have never been observed in combination with the current visual input; 

\textit{Predictions Related to One’s Own Actions or External Events:} In the egocentric scenario, it is relevant to distinguish between predictions depending on the wearer's own activities and predictions depending on changes of the environment. For example, in a pedestrian trajectory forecasting task, it is possible to model the future trajectories of the camera wearer based exclusively on the human's past trajectory without taking into account the environment, or analyse the context and model social interactions between pedestrians;

\textit{Internal State Predictions:} Rather than predicting future outputs directly, it may be beneficial for a system to predict its own internal future state (e.g., in the form of a future representation). It is also worth mentioning that egocentric future prediction may also be useful as a supplementary task to perform high-quality detection and recognition tasks in an online (streaming) scenario \citep{li2020towards}.

{By the time of writing, different survey papers overviewing the egocentric vision domain have been published. \cite{betancourt2015evolution} gave a comprehensive summary of the evolution of first-person vision from the very beginning to 2015. The survey paper presented in \citep{bandini2020analysis} focused on the analysis of hands from the egocentric vision perspective. Egocentric activity recognition in life-logging datasets has been reviewed in \citep{hamid2017survey}. The authors of \citep{del2016summarization} have studied summarization of egocentric videos. While all of the above-mentioned surveys analyse papers on egocentric vision, none of them set their focus on tasks concerning future prediction.}

{Other surveys do not consider egocentric vision explicitly \citep{katsini2020role, kong2018human, rasouli2020deep}, but are relevant to the topics of this survey. For example, eye gaze and its role in security applications is analysed in \citep{katsini2020role}, whereas multiple methods for action recognition and anticipation are listed in \citep{kong2018human}. A general overview of deep learning approaches for future prediction has been given in \citep{rasouli2020deep}. It should be noted that while the survey of \citep{rasouli2020deep} discusses a few methods considering egocentric scenes, its main focus is on third-person vision.}

{While previous surveys have considered the topic of future predictions from egocentric vision only marginally, this survey aims to offer a complete overview, reviewing the main challenges and applications in the domain of egocentric future predictions, and provide a comprehensive account of egocentric datasets, devices used for their collection, as well as learning architectures designed to predict the future from egocentric video.}

{To the best of our knowledge, this is the first survey aggregating research in the context of future prediction from egocentric video.}

Our objective is to focus on common problems, highlight the most effective techniques for solving future prediction tasks, and highlight areas that need further investigation by the research community. The bibliography collected in this survey is mostly from the egocentric vision domain, however, other papers on machine learning, computer vision and psychology are listed to provide a full-view analysis of the field.

The manuscript is organised as follows: in Section 2, we discuss real-life applications of egocentric future predictions methods; Section 3 reviews the devices which can be used for collecting egocentric data and implement egocentric future predictive systems; Section 4 introduces the challenges related to egocentric future prediction; in Section 5, we list and compare the main datasets commonly used in the literature; Section 6 reviews methods for egocentric future prediction, considering the input data modalities and model components; finally, in Section 7, we summarise the current state of research on egocentric future prediction and discuss possible directions for further research.

\section{Application Areas}

{In this section, we discuss the main application scenarios for technologies able to predict future events from egocentric video.}

{It should be considered that the field of egocentric future prediction is very young and still developing. Thus, some applications might be still only potential at the current stage. Nevertheless, we think it is still relevant to discuss about scenarios in which future prediction methods are likely to be useful, for example, to avoid the user to touch dangerous objects.}

{We believe, that this survey will help in identifying the possible application areas.}


\textit{Assistive technologies in daily living or industrial scenarios.} {Future predictions from egocentric videos can be beneficial for assistive technologies either in daily living or industrial scenarios \citep{leo2017computer}. In \citep{ohn2018personalized}, egocentric trajectory forecasting is used to understand a blind person's intentions and guide them avoiding collisions. In the industrial domain, future prediction is a natural need. Anticipating activities performed by a worker is important to check the correctness of the performed tasks and to prevent dangerous activities. For example, industries are recently supporting research projects aimed at developing solution for smart assistance in industrial scenarios. An example of such projects is ENIGMA, which has STMicroelectronics among its stakeholders. The project aims to predict and prevent potential risks of workers in an industrial scenario\footnote{https://iplab.dmi.unict.it/ENIGMA/}. Despite this interest, data acquisition in industrial scenario is problematic because of privacy concerns and the need of industrial secrets protection. Hence, currently there is no publicly available dataset of egocentric videos in real industrial scenarios. A recent example of a dataset of egocentric videos collected in an industrial-like scenario present by \citep{ragusa2020meccano}.} %


\textit{Human-robot interaction.} Human-robot interaction is important for many applications. {One possible solution to model human-robot interaction and navigating the robot consists in using wearable devices aimed at providing user-related information (i.e., user location) to a robot. There are examples in the literature showing that bracelets with sensors can be used for this purpose \citep{scheggi2014cooperative}. We believe that an interesting direction is to utilize smart glasses for providing the robot with enhanced information about user's location and performed activities.}
Anticipating where the person will go in the near future and what they will do can be useful for a robot to avoid crashing into a person and providing timely assistance.
Such tasks require fast and precise processing of video data, and egocentric cameras installed on a robot or being weared by a human could provide an advantageous view to capture the most important aspects of the performed activities \citep{ryoo2015robot, koppula2015anticipating}. For human-robot interaction, trajectory forecasting is one of the key research tasks \citep{manglik2019forecasting}.

\textit{Entertainment.} Augmented and virtual reality are becoming increasingly important in the development of the gaming and entertainment industries \citep{cacho2020emerging, hartmann2020entertainment}. AR- or VR-powered games provide the opportunity to feel a full-body experience especially in dance and sports scenarios. The diving into gameplay provided by means of AR and VR technologies requires to process data from first-person wearable devices and can benefit from action recognition and anticipation from egocentric view. Indeed, being able to anticipate where the human will look next and what virtual objects they will interact with can be useful to improve rendering and developing more natural user interfaces. Current works building such entertainment systems are focused on the recognition and prediction of hand gestures and body movements \citep{liang2015ar, taylor2020towards}. Even if these works are mainly focusing on recognition, methods dealing with future predictions may be beneficial in this domain.

\textit{Autonomous Driving Cars.} Although a survey of approaches specifically designed to anticipate future events in the context of self-driving cars is outside the scope of this paper, ideas and models developed to predict the future from first person videos can be useful in this domain. 
Indeed, videos collected form a self-driving car are based on an egocentric view of the scene, similarly to first person videos collected by humans.
The main tasks useful in this area include
either pedestrian intents prediction \citep{marchetti2020multiple, liu2020spatiotemporal, poibrenski2020m2p3}, future vehicle localization \citep{yao2019egocentric}, or both \citep{li2019learning}. Another important task in this setting, is predicting the time to the occurence of an event, such as a stopping or a collision \citep{neumann2019future}. A comprehensive overview of models for future prediction for autonomous vehicles is presented in \citep{hirakawa2018survey}.

\section{Devices}

%

\begin{table*}[!htpb]\centering
\caption{
Summary table of devices for egocentric future predictions.
}\label{tab:devices}
\scriptsize

\begin{tabular}{|C{2.6cm}|C{2.4cm}|C{1.8cm}|C{2.2cm}|C{1.2cm}|C{1.2cm}|C{1.8cm}|}
\hline
Device &Resolution &Field of View &Datasets  &Additional sensors& Computing Capabilities &High-level functionalities\\ \hline
SMI ETG 2w &1280x960p @24 fps; 1024x720p @30 fps; &80$^{\circ}$ horizontal, 60$^{\circ}$ vertical &EGTEA GAZE+ GTEA Gaze+ EPIC-Tent* & Gaze &no & -\\ \hline
ASL Mobile Eye XG &1600 x 1200 @ 60 fps &36$^{\circ}$, 60$^{\circ}$, 90$^{\circ}$ &BEOID  & Gaze &no & -\\ \hline
Narrative Clip &5 mP @ 2 fpm &70$^{\circ}$ &EDUB-SegDesc  & GPS &no &photostream segmentation \\ \hline
Point Grey FL2-08S2C-C &1032 x 776 @ 30fps &90$^{\circ}$ horizontal, 35$^{\circ}$ vertical &CMU  & - &no & -\\ \hline
Intel RealSense {(SR300, R200)} &1920x1080 @ 30fps &73$^{\circ}$ horizontal, 59$^{\circ}$ vertical &EgoGesture, MECCANO, {DoMSEV}  & Depth &no &SLAM\\ \hline
GoPro Hero 7 &2.7K @ 120 fps &wide &EK100  & IMU, GPS &no & - \\ \hline
GoPro Hero 5 &3840x2160 @ 60 fps &wide &EK55, EPIC-Tent
& IMU, GPS &no & -\\ \hline
GoPro Hero3+ &1920 x 1080 @ 30 fps 1280 x 720 @ 60 fps &wide &HUJI EgoSeg, ADL, {DoMSEV}
& - &no & -\\ \hline
Google Glass &720p @ 30 fps &narrow &KrishnaCam, EgoHands
& Gaze &yes &AR\\ \hline

Microsoft HoloLens 2&1080p @ 30 fps & 52$^{\circ}$ & EGO-CH  & Gaze, depth &yes &AR/MR, localization, hand tracking, SLAM\\\hline

Vuzix Blade & 720p @ 30fps 1080p @ 24fps & 60$^{\circ}$ & - &IMU, GPS &yes &AR, voice control\\\hline

Epson moverio BT-300 &30fps @ 720p &23$^{\circ}$ & -  & IMU, GPS & yes & AR\\\hline

Magic Leap & 1280x960 @ 120fps &50$^{\circ}$ & -  & Gaze, IMU, GPS & yes & AR/MR, voice control, hand tracking\\ \hline

Nreal Light & 1080p @ 60fps &52$^{\circ}$ & -  & IMU, GPS & yes & hand tracking, AR/MR, SLAM\\ \hline

Pupil Invisible & 1088x1080 @ 30fps & 82$^{\circ}$x82$^{\circ}$ & -  & Gaze, IMU & yes & NNs for gaze estimation\\ \hline

\end{tabular}
\end{table*} %

Modern first-person vision devices can serve as cameras for data collection, but they also offer other functionalities such as computing capabilities, eye tracking, localization and augmented reality, which makes them useful for implementing assistive systems based on future prediction capabilities. 
Table \ref{tab:devices} lists the main devices, with a particular focus on those which have been used for the collection of datasets useful for egocentric future predictions.
For each device, we list resolution, field of view, datasets collected using the device, additional sensors apart from cameras, whether the device has computing capabilities, and possible high-level functionalities provided by the device.
Note that feature-rich devices including additional sensors (e.g., to collect gaze, depth, or to estimate location), are of particular interest for future predictions. 
We divide devices
into 6 categories, as discussed in the following.

\textit{Action cameras} are generally used to capture sports videos or to perform video-based life-logging. The most popular devices from this group are GoPro cameras\footnote{https://gopro.com/}, which have been used to collect many datasets of egocentric videos~\citep{damen2020rescaling,damen2018scaling,jang2019epic,pirsiavash2012detecting}. Action cameras allow to capture high quality egocentric videos which are suitable to study future predictions as in the case of the EPIC-KITCHENS dataset series~\citep{damen2018scaling,damen2020rescaling}. Among the listed device categories, action cameras have the lowest price, due to the wide availability as consumer products on the market. 

\textit{Life-logging cameras} are designed to take photos of everyday activities within a long time span (e.g., one entire day). An example of life-logging camera is the Narrative clip\footnote{http://getnarrative.com/}, designed to capture a sequence of images at a very low frame rate (2 frames per minute), thus allowing to create daily collections of life-logging photos. Previous examples of such devices, albeit now discontinued were the Autographer\footnote{https://en.wikipedia.org/wiki/Autographer}, which could capture up to 2000 pictures a day, and Microsoft SenseCam\footnote{https://www.microsoft.com/en-us/research/project/sensecam/}, with similar capabilities, and mainly designed as a research tool. In \citep{bolanos2018egocentric}, Narrative Clip was used to collect a dataset of life-logging activities used for action segmentation. Even if this research topic has not yet been thoroughly investigated, this huge amount of life-logging data could be used to predict the user future activities, by modelling regular behavioural patterns.

\textit{Eye trackers} allow to capture egocentric video and measure the spatial locations in the frame the user is looking at. Popular eye-trackers are the SMI eye-tracker\footnote{https://imotions.com/hardware/smi-eye-tracking-glasses/}, the ASL Mobile eye and the Tobii eye-tracker\footnote{https://www.tobii.com/}. Pupil Invisible\footnote{https://pupil-labs.com/} utilizes built-in computational resources to model human gaze with neural networks. Eye-trackers may also be implemented as an additional functionality in smart glasses. For example, a gaze tracker is available in Microsoft Hololens 2\footnote{https://www.microsoft.com/it-it/hololens/}. Eye tracking data is especially important in the future prediction scenario. Indeed, gaze can serve as a clue on where a humans point their attention, so it may be an indicator for what they will do next~\citep{land2006eye}. 

\textit{Smart glasses} are generally designed as pair of regular glasses including different sensors such as one or more cameras, a microphone, GPS-receivers, accelerometers and gyroscopes. Most designs include on-board computing capabilities or the ability to communicate with a smartphone, as well as a mechanism to provide feedback to the user such as a head-up display or a bone-conduction audio speaker.
Given their design, these devices can be naturally operated in a hands-free fashion, which provides unique possibilities for the development of natural interfaces for user-device interaction.
Smart glasses can be exploited for assistive tasks \citep{mcnaney2014exploring, dougherty2017using}, which are likely to benefit from future predictions capabilities. 
The most popular devices on the market are 
Google Glass\footnote{https://www.google.com/glass} and Vusix Blade\footnote{https://www.vuzix.com/products/blade-smart-glasses-upgraded}. Despite their potential, smart glasses are still not adopted by consumers on a daily basis, with the most popular areas of application being assistive manufacturing\footnote{https://www.vuzix.com/solutions/manufacturing} and tele-medicine\footnote{https://www.vuzix.com/Solutions/telemedicine}.
It is likely that the smart glasses market will grow as new use cases for assistive applications in a variety of scenarios ranging from daily activities to the industrial sector including predicting abilities are investigated.

\textit{AR/MR glasses}. Augmented and mixed reality devices can allocate generated virtual objects within the scene captured from the egocentric perspective. These abilities allow to design more natural user interfaces which can be beneficial for assistive systems. Hence, the development of AR technologies serves as a boost for egocentric vision research. 
Differently from smart glasses, AR/MR glasses generally include a high quality display with a wide Field of View, dedicated graphics processing units, as well as multiple sensors to track the position of the device in the 3D space. The most noticeable devices in this category are Microsoft Hololens 2 and Magic Leap \footnote{https://www.magicleap.com/}. 
A more recently introduced device in this category is Nreal Light\footnote{https://www.nreal.ai/}, the first egocentric vision device supporting 5G technology.  AR glasses can be useful in many application scenarios. However their current prices tend to be high, so they are mostly employed for research and industrial applications, rather than as devices for personal use.

\begin{figure*}[h]
\centering
\includegraphics[width=0.8\textwidth]{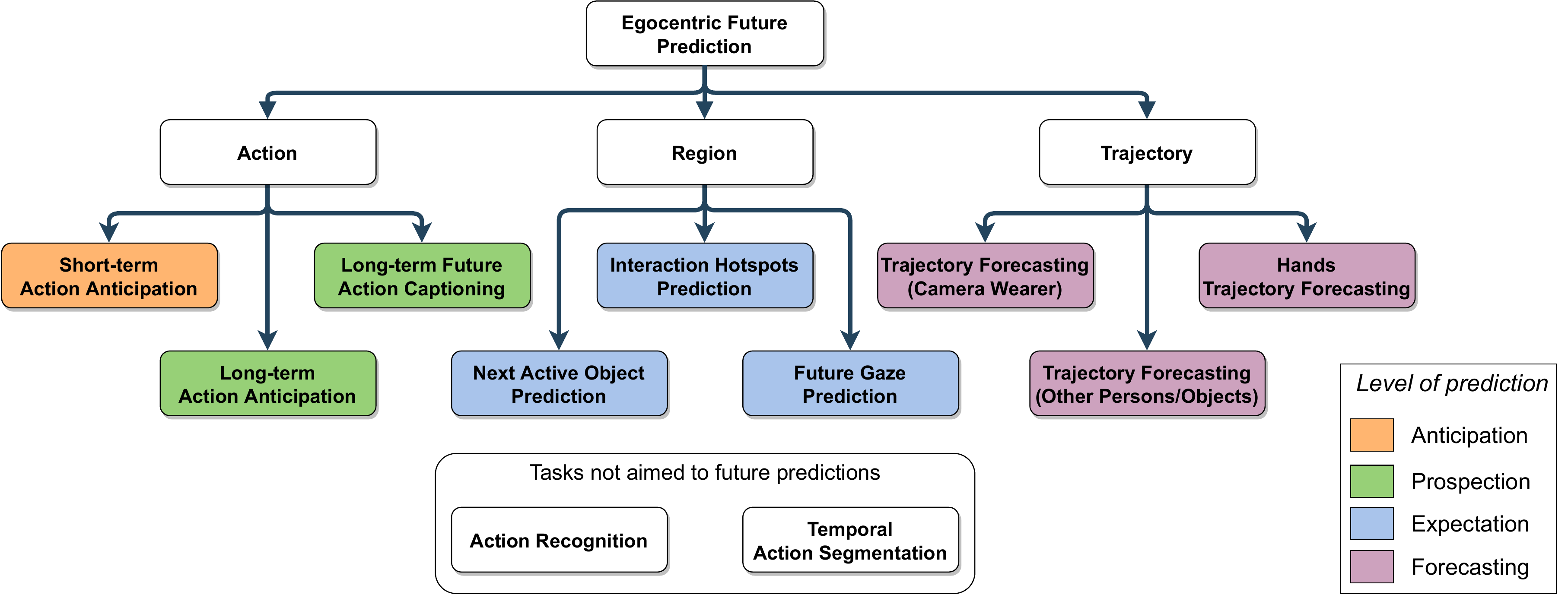}
\caption{Objectives of egocentric video processing for future prediction in the state-of-the-art literature.}
\label{fig:objectives}
\end{figure*}

\textit{Other devices}. Although not widely used in everyday life, other devices besides the aforementioned categories can be useful for applications and research on future prediction in egocentric vision. For example, during the collection of the NearCollision dataset~\citep{manglik2019forecasting} a LIDAR sensor was used for capturing scene depth. The CMU dataset \citep{de2009guide} captured human-object interactions by using RF-ID tags. RGB-D cameras may also be used in the considered scenario, but to the best of our knowledge, currently there is no works which uses depth for making predictions about the future. 
Another important type of devices to mention is event cameras, such as the dynamic and active-pixel vision sensor (DAVIS) \citep{patrick2008128x} which can be used for computer vision tasks such as SLAM or pose estimation \citep{mueggler2017event}. Despite their potential, event cameras are not yet widely used for egocentric vision. 
It is worth noting that recent wearable devices are beginning to include have more sensors apart from cameras. For example, Microsoft Hololens 2 has a RGB camera, several grayscale cameras used for position tracking, a depth camera, an eye-tracker, as well as accelerometers and gyroscopes.

\section{Challenges} 
\label{sec:challenges}
We group the main challenges related to future prediction in egocentric vision into three main clusters and nine different challenges, as reported in Figure~\ref{fig:objectives}.
We also consider $2$ additional challenges such as temporal action segmentation and action recognition which are not oriented to future predictions but are closely related to the investigated ones.
We adopt the terminology introduced in Table~\ref{tab:definitions} and assign a ``level of prediciton'' to each challenge.
In particular, we put challenges aiming to predict abstract future events such as action labels in the ``anticipation'' level if they are short-term and in the ``prospection'' level if they are long-term. We classify trajectory-related tasks with ``forecasting'' as they deal with time-series data in both their inputs and outputs.
Finally, we classify challenges aiming to predict future interacted objects or image regions with ``expectation'', since their predictions are grounded into the visual appearance of the scene.
The identified challenges are discussed in the following.


\subsection{Action prediction}
The challenges in this cluster aim to predict the future actions, either in the form of labels or captions, which will be performed by the camera wearer in the short- or long-term future.

{\textit{Short-term Action anticipation.} Short-term action anticipation (often referred to simply as ``action anticipation'') is the task of predicting an action before it occurs. The action that will start at timestamp $\tau_{as}$ (action start) is to be predicted from the observation of a video segment preceding the action start time by an anticipation time $\alpha$. Thus, considering fixed segment length $l$ and the anticipation time $\alpha$, the processed video segment is bounded by starting time $\tau_{as} - l - \alpha$  and ending time $\tau_{as} - \alpha$. Actions can be modelled with a single label, or with a (verb, noun) pair, e.g., \textit{``pour water'', ``clap hands''}.} Prior to the introduction of the EPIC-KITCHENS-55 benchmark (EK-55) for training and evaluating egocentric action anticipation models in 2018 \citep{damen2018scaling}, works on action anticipation mainly considered third-person scenario \citep{vondrick2016anticipating, gao2017red, rodriguez2018action}. Besides EK-55 dataset, its extension EK-100 \citep{damen2020rescaling}, and EGTEA Gaze + \citep{tech2018extended} datasets are suitable for training action anticipation models. The EPIC-KITCHENS action anticipation challenge introduced in \citep{damen2018scaling} allowed to unify the task by providing a benchmark, which led to a growth in the number of publications focusing on egocentric future anticipation in recent years \citep{furnari2019would, camporese2020knowledge, liu2019forecasting, miech2019leveraging, zhang2020egocentric,sener2020temporal}.

\textit{Long-term action anticipation.} The task of long-term anticipation consists in predicting the sequence of next actions performed by the camera wearer. In the literature, this task may also be referred to as ``dense anticipation'' \citep{sener2020temporal}. 
Some works~\citep{abu2018will, mahmud2017joint} perform experiments on datasets of instructional activities~\citep{kuehne2014language, stein2013combining} which often resemble egocentric videos due to the framing focusing on the hands.
Other works~\citep{sener2020temporal, ke2019time, zhaodiverse} perform validation on datasets of natural activities, such as EPIC-KITCHENS. By the time of writing there is no common benchmark for evaluating the performance of models designed for egocentric long-term future anticipation.

\textit{Long-term future action captioning.} While in the above mentioned challenges actions are described with a single label, or with a (verb, noun) pair, the task of future action captioning consists in generating natural language text describing future activities (e.g., ``add the caramel sauce and stir until the chocolate is melted'' ). 
Compared to labels, textual descriptions may provide more detailed information about actions which are going to take place. Hence learning how to represent the predictions about future actions as text can be an promising research direction. Currently few egocentric datasets provide text annotations (e.g., EPIC-KITCHENS~\citep{damen2018scaling} and \citep{bolanos2018egocentric} provide textual narrations of actions), so only few works focus on future action captioning from the egocentric perspective, and none of them deals with non-scripted activities. \citep{sener2019zero} explored the possibility to predict the next steps of cooking activities in the form of a recipe. While future action captioning may be both long-term (predicting multiple or distant events), or short-term (predicting an action that is going to occur shortly after the observation), we have not found works in the literature focusing on short-term future captioning, although it may be a possible new challenge. For the simplicity, later in text we will refer long-term future action captioning simply as ``future action captioning''.

\subsection{Region prediction}
The challenges in this cluster aim to predict which regions of the frame contain elements the camera wearer will focus on in the future. These regions can be represented as object bounding boxes, interaction hotsposts or future gaze locations.

\textit{Next active object prediction}. Predicting next active objects consists in detecting the objects in the scene which will be used by the camera wearer as part of a human-object interaction \citep{furnari2017next}. Although actions may involve more than one object, (e.g., cutting onion involves \textit{knife} and \textit{onion}), predicting several next active objects at once is not widely represented in the egocentric vision research. 
Related lines of works have considered the problems of
\textit{detecting action-objects} (i.e., objects which will not necessary be involved in future actions, but attract the camera wearer's attention)~\citep{bertasius2016first,bertasius2017unsupervised} and predicting the location at which objects and hands will appear in future frames (e.g., 1 or 5 seconds later)~\citep{fan2018forecasting}.
Datasets for future active object prediction should contain untrimmed videos with active object annotated both during the interaction and before its beginning.
Examples of such datasets include ADL~\citep{pirsiavash2012detecting} and EPIC-KITCHENS-55~\citep{damen2020rescaling}.
Although there are a few datasets suitable for future object modelling, currently there is no standard benchmark for training and evaluating models aimed to future object prediction, which has likely limited its investigation in past works.

\textit{Predicting interaction hotspots.} 
Other works have considered the problem of
\textit{predicting interaction hotspots} \citep{liu2019forecasting, nagarajan2019grounded}, which are defined as regions in a frame 
which the camera wearer will likely touch in the near future. 
Examples of interaction hotspots are objects parts such as a pan's handle or objects such as a knife.
Although only few works have considered this task, the ability to detect interaction hotsposts can be beneficial for applications related to safety.

\textit{Future gaze prediction.} Human gaze can provide useful information on the camera wearer's attention and their intentions. Gaze fixation points often indicate the object the camera wearer is going to interact with, or highlight the area in which the object is located~\citep{land2006eye}. Although gaze can be an important modality to anticipate future actions, only few works have focused on anticipating gaze itself \citep{zhang2017deep, aakur2020unsupervised}.

\subsection{Trajectory prediction}
The challenges in this cluster aim to predict future trajectories either of the camera wearer, of their hands, or of other persons.

\textit{Forecasting the trajectory of the camera wearer} aims to predict a plausible sequence of future coordinates of the camera-wearer $(x_{t+1},y_{t+1}),..,(x_{t+n}, y_{t+n})$ starting from the observation of a video clip ending at timestep $t$ and an initial point with coordinates $(x_t, y_t)$. 
If only the last point $(x_{t+n}, y_{t+n})$ is required, the task may be referred to as \emph{future localization} \citep{soo2016egocentric}. 
Predicting plausible future trajectories of ego-motion in egocentric videos requires to learn to avoid obstacles and move between objects and humans. 
Previous works have tackled this challenge on different datasets considering walking~\citep{soo2016egocentric,singh2016krishnacam} and sports scenarios~\citep{bertasius2018egocentric}.
Unfortunately, the lack of a standard benchmark to address the task has limited the development of approaches despite the potential technological impact.
Indeed, predicting plausible future trajectories in the scene can be beneficial in assistive navigation technologies \citep{zhao2019designing} as well as in robotics \citep{soo2016egocentric}.

\textit{Forecasting the trajectories of other persons or objects.} This challenge consists in determining the future trajectories of other people (e.g., pedestrians) or objects (e.g., vehicles) from an egocentric perspective.
The tasks of predicting human's trajectories from video has been extensively studied in the context of third person vision~\citep{alahi2016social, gupta2018social, kothari2020human,giuliari2020transformer}. In particular, the TrajNet challenge~\citep{sadeghian2018trajnet} has recently encouraged the investigation of several models for human trajectory forecasting which mainly assumed a surveillance scenario.
Another related line of works has focused on predicting pedestrians' and vehicles' trajectories from the egocentric point of view of autonomous vehicles~\citep{neumann2019future, liu2020spatiotemporal, malla2019nemo, poibrenski2020m2p3,Marchetti_2020_CVPR}.
Despite these attempts, this challenge has not been extensively studied in the first person vision scenario in which the camera is worn by a human.
Indeed, the only investigation in this scenario is the one of \citep{yagi2018future}, which focused on forecasting other persons' trajectories from egocentric video. 
One of the main limitations for the development of this challenge is the lack of benchmark
datasets with dense annotations of people's locations. 
For instance, the authors of \citep{yagi2018future} evaluated their approach on their own private dataset of only about 4.5 hours of recordings.

\textit{Hands trajectory forecasting.} 
Hands play a fundamental role in the modelling of human-object interactions from an egocentric perspective~\citep{shan2020understanding}.
While previous works have investigated the role of hands and their trajectories to model tasks such as action recognition~\citep{fathi2011learning,li2015delving,ma2016going,kapidis2019egocentric}, their usefulness for future predictions has not yet been systematically investigated in the literature.
The authors of~\citep{liu2019forecasting} proposed a method that implicitly predicts ``motor attention'' -- the future trajectory of the hands to model interaction hotspots -- and predict future actions, which suggests the importance of hands trajectories for future predictions.
Also in this case, we would expect the emergence of suitable datasets containing hand track annotations and standard benchmarks to encourage research in this challenge.

\subsection{Related challenges}

In this subsection, we discuss additional challenges which do not directly address future prediction tasks, but are highly related to the topic. Indeed, previous works on future predictions have generally considered models developed for these tasks as off-the-shelf components, or as baseline approaches. For example, models designed for action recognition have shown reasonable baseline results for action anticipation \citep{damen2018scaling}, or have been used to extract features for action anticipation \citep{furnari2019would,sener2020temporal}.

\textit{Action recognition}. From a given input sequence of temporal bounds $\tau_{as}...\tau_{as+ad}$ where $as$ and $ad$ are the action start timestamp and action duration respectively, the task is to classify the action $y$ captured on the sequence. Although this task is not about modelling and predicting the future, most of the methods developed for recognition can be exploited as baselines for anticipation. For example, Temporal Segment Network (TSN) \citep{wang2016temporal} was employed as a baseline for both the action anticipation and action recognition on the EPIC-KITCHENS dataset in \citep{damen2018scaling}. Also, previous works on action anticipation~\citep{furnari2019would,sener2020temporal} have relied on TSN-based features.

{\textit{Temporal Action segmentation} (also referred to as temporal action localization or temporal action detection) consists in detecting all temporal segments containing the action happening in a video~\citep{lea2017temporal,farha2019ms,ma2020sf,li2020ms}.
The task of video segmentation has been thoroughly explored also in egocentric vision to highlight actions~\citep{spriggs2009temporal}, long-term activities~\citep{poleg_cvpr14_egoseg}, and locations~\citep{furnari2018personal}.
Although action segmentation is not directly focused on future predictions, extracted action segments from the past have been used to forecast the future action sequences and their duration \citep{sener2020temporal, abu2018will}.}



\section{Datasets}

This section discusses the main datasets of egocentric videos which are relevant to research on future prediction tasks. The datasets vary in size, level of provided annotations, captured environments and tasks. A detailed list of 20 datasets categorised by 17 attributes relevant to future prediction tasks is discussed in the following section and presented in Table \ref{tab:datasets}.










\subsection{Attributes}
\textit{Task.} The challenges introduced with the dataset allows to understand the applicability of data to various research purposes. Although several datasets were not initially designed for challenges aimed at future predictions, technically, most of the datasets can be adapted to study them. For example, dataset designed for action recognition may be used for action anticipation if they contain long untrimmed videos, which allows to sample frames from the video prior to the beginning of the action. Most of these datasets can be used as starting points to add labels specific to future prediction tasks.
    
\textit{Scripted/Non-scripted.} While in scripted datasets subjects are asked to perform specific activities according to a script (e.g., ``take item A, then go to place X''), non-scripted datasets capture activities at their natural speed or complexity.
Non scripted datasets are more suitable for egocentric future prediction, since algorithms trained on that type of data will tend to learn natural behavioural patterns, rather than simplistic scripted sequence of activities.
    
\textit{Hours of video/Number of frames.} The number of hours of video is one of the main characteristics when it comes to evaluating dataset size. However, various datasets are captured with different frame rates. Thus depicting the number of video frames is also important to better understand dataset size.

\textit{Videos.} The number of video recordings in the dataset is another important parameter for research on future prediction. Indeed, different videos may include different scenarios (e.g., different scenes or different activities) a predictive system should be able to work on.
    
\textit{Segments.} For action recognition datasets, the number of segments indicates the number of samples related to the action classes and action beginning/ending timestamps, whereas for trajectory forecasting datasets it indicates the number of coordinate sequences. The number of segments often correspond to the number of training examples in the case of fully-supervised tasks (e.g., action anticipation), hence a large number of sequences is an important attribute to consider. 
    
\textit{Classes.} This attribute defines the number of classes in the dataset, which is relevant to assess the complexity of the semantic annotations. 
For action anticipation/recognition, it reports the number of action classes, whereas for next active object prediction, it reports the number of object classes. This attribute does not apply to trajectory forecasting datasets. 
    
\textit{Native environments.} Whether videos have been collected in environments which are native for the subjects (e.g., their homes) or not (e.g., laboratory environments prepared specifically for the collection of the dataset). Native environments are preferable for research on future predictions as subjects tend to behave more naturally in familiar places.
    
\textit{Number of environments.} The number of different environments considered during the data collection. The more environments are captured on a dataset, the more scalable to new environment are the models trained on that dataset likely to be.
    
\textit{Subjects.} The number of different subjects recording the videos. As with environments, a large number of subjects is an important characteristic which may allow to model general human behaviour patterns.
    
\textit{Object/Hand annotations}. Whether the dataset contains object/hand annotations. The importance of processing hands and objects annotations is discussed in detail in Section 
\ref{modalities}.
    
\textit{Audio}. Whether the videos include audio. As discussed in Section \ref{modalities}, audio can be a useful modality for future predictions.
    
\textit{Scenario.} The general scenario in which the recordings took place. This attribute allows to understand the domain of applicability of models trained on the considered data.
    
\textit{Number of persons.} Number of persons involved in a single video. Most datasets do not consider a multi-person setting, and the camera wearer is often the only person in the scene. This parameter allows to understand whether a dataset is suitable for studying social interactions or not.
    
\textit{Trimmed.} Whether the videos are trimmed (e.g., short videos with specific activities performed by the subject), or untrimmed, with annotations of starting and ending timestamps of the events of interest (if relevant). Untrimmed videos are more useful to study future prediction tasks since they allow to use parts of the video which are captured before the beginning of the event to be predicted. Untrimmed videos containing long activities are also important for learning dependencies between subsequent actions.
    
\textit{Camera position.} {How the subject wears the first-person camera. Possible camera positions include head or chest (\textbf{H} and \textbf{C} in Table~\ref{tab:datasets} respectively), but other options are also possible, such as, suitcase (\textbf{S}), or systems in which two cameras are mounted on head and hand (\textbf{HH}). According to \cite{mayol2009choice} head mounted cameras allow to model the wearer's attention and provide larger field of view comparing to chest- or hand-mounted cameras. However, head-mounted cameras are more exposed to head and body movements, which should be considered when modelling egocentric vision systems. Thus, the camera position should be aligned to the specific application context the algorithm will be deployed to.}



\begin{table*}[!htp]
\centering
\caption{List of datasets of egocentric videos (see text for column descriptions). The following abbreviations are used in the table. Task: Action Recognition (AR), Action Anticipation (AA), Action Captioning (AC), Object Detection (OD), Next Active Object (NAO), Gaze Prediction (GP), Trajectory Forecasting (TF), Action Segmentation (AS),
Time to Action (TTA), Pose Estimation (PE), Semantic Segmentation (SS), Object Search (OS), Uncertainty Estimation (UE),
Domain Adaptation (DA), Video Generation (VG). Camera Position: Mounted on head (H), Mounted on head and hand (HH), Mounted on suitcase (S), Mounted on Chest(C), Mounted on Shoulder (Sh).
Duration/Frames: The ``$^\ast$'' symbol indicates that we considered only the length of the egocentric part of dataset.
Not Provided (N/P) and Not Applicable (N/A) indicate whether a given attribute has not specified by the authors or is not applicable.
}

\adjustbox{max width=\textwidth}{
\setlength{\tabcolsep}{2pt}
\scriptsize
\begin{tabular}{|L{1.8cm}|C{1.2cm}|C{0.8cm}|C{0.8cm}|C{1.1cm}|C{0.95cm}|C{1cm}|C{1cm}|C{1cm}|C{1cm}|C{0.9cm}|C{1cm}|C{1cm}|C{1.1cm}|C{0.8cm}|C{0.9cm}|C{0.9cm}|C{0.9cm}|C{0.7cm}} 
\hline
\textbf{Paper}  &\textbf{Task} &\textbf{Unscripted} &\textbf{Duration} &\textbf{Frames} &\textbf{Videos} &\textbf{Segments} &\textbf{Classes} &\textbf{Native environments} &\textbf{\# of envs} &\textbf{Subjects} &\textbf{Object annotations} &\textbf{Hand annotions} &\textbf{Scenarios} &\textbf{\# of persons} &\textbf{Audio} &\textbf{Untrim-med} &\textbf{Camera Position} \\
\hline

%

CMU-MMAC \citep{de2009guide} &AR, PE, anomaly &no &N/P &0.2M &16 &516 &31 &no &1 &16 &RFID tags &motion capture &indoor &1 &yes &yes &H \\ \hline

ADL \citep{pirsiavash2012detecting} &AR, NAO &no &10h &1.0M &20 &436 &32 &yes &20 &20 &137.780 bboxes &bboxes &indoor, home &1 &yes &yes &H \\ \hline

GTEA Gaze+ \citep{fathi2012learning} &OD, AR, GP &no &~10h &0.4M &35 &3.371 &42 &no &1 &10 &no &yes &kitchen &1 &yes &no &H \\ \hline

HUJI EgoSeg \citep{poleg_cvpr14_egoseg} &AR, GR &no &65h &N/P &122 &122 &14 &no &N/P &14 &no &no &indoor \& outdoor &3+ &no &no &H \\ \hline

BEOID \citep{damen2014you} &NAO, AR, GP &no &1h &0.1M &58 &742 &34 &no &6 &5 &yes &no &kitchen, office, gym &1 &no &no &H \\ \hline

EgoHands \citep{bambach2015lending} &OD, SS, AR &yes &1.2h &130K &48 &48 &4 &yes &3 &4 &no &15.053 masks &indoor \& outdoor &2 &yes &yes &H \\ \hline

KrishnaCam \citep{singh2016krishnacam} &TF &yes &70h &7.6 M &460 &N/A &460 &yes &N/P &1 &no &no &outdoor &3+ &no &yes &H \\ \hline

OST \citep{zhang2017deep} &OS &no &14h &0.5M &57 &57 &22 &yes &1 &55 &yes &no &indoor &1 &no &yes &H \\ \hline

Charades-ego \citep{sigurdsson2018charades} &AR, AC &no &34.4h$^\ast$ &2.3M &2.751 &30.516 &157 &yes &112 apartments, 15 locations &112 &no &no &flat &1 &no &no &HH \\ \hline

ATT \citep{zhang2018coarse} &GP, NAO &yes &2h &217K &2h &N/A &N/A &yes &N/P &N/P &no &no &indoor &2 &yes &N/P &H \\ \hline

EGTEA GAZE+ \citep{tech2018extended} &AR, GP &no &28h &2.4M &86 &10.325 &106 &no &1 &32 &no &15.176 masks &kitchen &1 &yes &yes &H \\ \hline

EgoGesture \citep{zhang2018egogesture} &AR &no &27h &3M &2.081 &24.161 &83 &yes &6 &50 &no &no &indoor \& outdoor &1 &no &no &H \\ \hline

EDUB-SegDesc \citep{bolanos2018egocentric} &AR, AC &yes &55d &48.717 &1.339 &3.991 &N/A &yes &N/P &11 &no &no &indoor \& outdoor &3+ &no &yes &H \\ \hline

ThirdToFirst \citep{elfeki2018third} &VG &no & 1h$^\ast$ & 109K$^\ast$ & 531$^\ast$ &531 &8 &no &4 &1 &no &no &outdoor &1 &no &no &H \\ \hline

DoMSEV \citep{silva2018weighted} & AR & yes & 8h & 0.9M & 73 & 73 & 13 & yes & 10 & N/P & no & no & indoor \& outdoor & N/P & yes & yes &H, C, Sh \\ \hline

EK-55 \citep{damen2018scaling} &AR, AA, OD &yes &55h &11.5M &432 &39.596 &2.747 &yes &32 &32 &454.225 bboxes &no &kitchen &1 &yes &yes &H \\ \hline

NearCollision \citep{manglik2019forecasting} &TTA &yes &N/P &N/P &13.685 &13.685 &1 &no &1 &N/P &no &no &indoor, office &3+ &no &no &S \\ \hline

EPIC-Tent \citep{jang2019epic} &AR, GP, UE, AS &no &5.4h &1.2M &24 &921 &12 &no &1 &24 &no &no &outdoor &1 &yes &yes &H \\ \hline

100DoH 
\citep{shan2020understanding} &OD, PE, NAO &yes &131d &N/P &27.3K &27.3K &N/A &yes &N/P &N/P &110.1K bboxes &186K bboxes &indoor \& outdoor &N/P &no &yes &N/P \\ \hline

EK-100 \citep{damen2020rescaling} &AR, AA, OD, DA  &yes &100h &20M &700 &89.979 &4.025 &yes &45 &37 &38M bboxes &31M bboxes &kitchen &1 &yes &yes &H \\ \hline

EGO-CH \citep{ragusa2020ego} & OS, OD &yes & 27h & 2.9M & 60 & 60 & 226 & yes & 26 & 70 & 177K frames w/ bboxes & no & indoor, museums & 1 & yes & yes &H \\ \hline

MECCANO \citep{ragusa2020meccano} & AR, NAO &no & 7h & 0.3M & 20 & 1.401 & 61 & yes & N/A & 20 & 64.349 bboxes & no & indoor & 1 & no & yes &H \\ \hline


\end{tabular}

}
\scriptsize

\label{tab:datasets}
\end{table*}

\subsection{Datasets}
In the following, we briefly revise the relevant datasets which are classified with respect to aforementioned attributes in Table \ref{tab:datasets}. Datasets are ordered with respect to the date of publication.

\textbf{CMU-MMAC} \citep{de2009guide}: Carnegie Mellon University Multimodal Activity Database consists of data acquired from multiple devices. Besides modalities widely represented in other datasets such as RGB frames, audio, and object annotations, the CMU-MMAC dataset provides motion capture, accelerometer, light sensor and third-person video data. these settings allow to model human behaviour from multiple perspectives.

\textbf{ADL} \citep{pirsiavash2012detecting}: Activities of Daily Living is a high-resolution dataset captured by subjects performing daily activities and provides action annotations, object and hands bounding boxes. Each object bounding box is marked as active (i.e., manipulated) or passive.

\textbf{GTEA Gaze +} \citep{fathi2012learning}: Egocentric dataset of subjects performing kitchen activities, consisting of about 10 hours of video recordings. It is among the few datasets including gaze annotations.

\textbf{HUJI EgoSeg} \citep{poleg_cvpr14_egoseg}: Dataset of trimmed videos with recorded indoor and outdoor activities. This dataset of high-resolution videos can be used for long term activity recognition (e.g., a person is walking, running, standing or riding a bike). 

\textbf{BEOID} \citep{damen2014you}: Dataset acquired by 5 subjects performing 34 activities. Annotations contain action labels, gaze point cloud, action annotations and 3D object map.

\textbf{EgoHands} \citep{bambach2015lending}: This dataset is focuses on analysing hands in egocentric videos. It consists of 48 first-person videos of people interacting in realistic environments, with over 15.000 pixel-level hand masks.

\textbf{KrishnaCam} \citep{singh2016krishnacam}: The largest dataset of egocentric videos for person's trajectory path forecasting. KrishnaCam consists of $460$ video sequences for a total $70$ hours. Annotations of the dataset include data from accelerometer, gyposcope and GPS sensors.

\textbf{OST} \citep{zhang2017deep}: A dataset for the object search task. It consists of 57 video sequences captured by 55 subjects asked to search for objects belonging to 22 classes. Eye-tracking annotations are available for all videos in the dataset.

\textbf{Charades-ego} \citep{sigurdsson2018charades}: A large-scale dataset of paired third and first person trimmed videos. At the moment of writing, it is the biggest dataset of paired videos captured in natural environments.

\textbf{ATT} \citep{zhang2018coarse}: Adults, Toddlers and Toys dataset comprises unscripted gaming activities performed by parents and their children simultaneously. ATT could be used for the analysis of human attention, its differentiation between children and adults, and their interaction behaviours.

\textbf{EGTEA Gaze +} \citep{tech2018extended}: The extended GTEA Gaze+ dataset contains 106 classes, 2.4M frames, 10.325 action segments and has been recorded by 32 participants. The dataset also provides hand masks and gaze annotations.

\textbf{EgoGesture} \citep{zhang2018egogesture}: A dataset designed for gesture recognition, consisting of 24.161 videos of 50 subjects performing 83 different gestures. Data for EgoGesture was collected with and RGB-D camera.

\textbf{EDUB-SegDesc} \citep{bolanos2018egocentric}: Contains frames sampled from 55 days of video. EDUB-SegDesc consists of 1.339 events with 3.991 short text activity descriptions (e.g., \textit{I entered in classrooms and talked to different people}), collected from 11 subjects. This is one of the few egocentric datasets that can be used for text generation.

\textbf{ThirdToFirst} \citep{elfeki2018third}: Paired dataset consisting of 531 pairs of egocentric and third-person videos from top and side views. The dataset is annotated with action labels. The challenge proposed with the dataset is to generate the first-person videos from the third-person input. The authors also provide a synthetic data with paired videos modelled in the unity 3D platform.

{
\textbf{DoMSEV} \citep{silva2018weighted}: An 80 hour dataset of natural activities performed in indoor and outdoor environments. DoMSEV provides multimodal data (RGB-D, IMU and GPS), as well as action class annotations. It was collected originally to create smooth fast-forward egocentric videos, without losing semantically meaningful parts. DoMSEV is one of the few egocentric datasets which provides depth modality.}

\textbf{EPIC-KITCHENS-55} \citep{damen2018scaling}: A large-scale high-resolution dataset with non-scripted activities, recorded by 32 participants in their native kitchen environments. The dataset counts 55 hours of video consisting of 11.5M frames, densely labelled for a total of 39.6K action segments and 454.3K object bounding boxes. Three challenges have been proposed with this dataset: egocentric action recognition, active object detection, and egocentric action anticipation. Actions are provided as verb-noun pairs with 125 verb classes and 331 noun classes, forming 2513 action classes.

\textbf{NearCollision} \citep{manglik2019forecasting}: a dataset designed to predict the time to ``near-collision'' between a suitcase-shaped robot pushed by its user and a nearby pedestrian. The camera is mounted on top of the suitcase, thus providing a unique point of view. Annotations contain stereo-depth and 3D point cloud. 

\textbf{EPIC-Tent} \citep{jang2019epic}: Outdoor video dataset annotated with action labels (tasks, e.g., put detail A into B), 
collected from 24 participants while assembling a camping tent. The dataset provides 2 views - one captured from a GoPro camera and another from an SMI eye-tracker. In addition to task classes annotations, EPIC-Tent comprises task errors, self-rated uncertainty and gaze position.

\textbf{100DoH} \citep{shan2020understanding}: a large-scale dataset (131 days of recordings) consisting of egocentric and exocentric videos from YouTube. 100DoH can be utilized for next active object prediction task. Since the dataset is collected from the internet, it provides the biggest number of actors and environments.

\textbf{EPIC-KITCHENS-100} \citep{damen2020rescaling}: Extended version of EPIC-KITCHENS-55, resulting in more action segments (+128\%), environments (+41\%) and hours (+84\%), with denser and more complete annotations of fine-grained actions. Also new challenges are introduced: action detection, weakly-supervised action recognition, multi-instance retrieval, as well as unsupervised domain adaptation for action recognition.

\textbf{EGO-CH} \citep{ragusa2020ego}: A dataset of egocentric videos for visitors’ behaviour understanding in cultural sites. The dataset has been collected in two cultural sites and the authors introduced four tasks related to visitor's behavior understanding: room-based localization, point of interest/object recognition, object retrieval and
visitor survey prediction from video.

\textbf{MECCANO} \citep{ragusa2020meccano}: Dataset of egocentric videos to study human-object interactions in industrial-like settings. MECCANO has been recorded by 20 participants interacting with small objects and tools to build a motorbike model. MECCANO can be used for several challenges: 1) action recognition, 2) active object detection, 3) active object recognition and 4) egocentric human-object interaction detection.

Several papers evaluate their future prediction methods not only in the egocentric scenario, but also on third-person vision data. Some relevant third-person datasets which have been used in the literature are:
Hollywood2 \citep{marszalek2009actions}, 50Salads \citep{stein2013combining}, Breakfast \citep{kuehne2014language}, ActivityNet-200 \citep{caba2015activitynet}, OPRA \citep{fang2018demo2vec}. 
Other datasets of egocentirc videos have been seldom used for evaluation and hence they are not 
listed in Table, \ref{tab:datasets} due to their small size or because they are not publicly available. Some examples of these datasets are: EgoMotion dataset \citep{soo2016egocentric}, Daily Intentions Dataset \citep{wu2017anticipating}, First-person Locomotion \citep{yagi2018future}, Stanford-ECM \citep{nakamura2017jointly}, Stanford 2D-3D-S \citep{armeni2017joint}.

\subsection{Discussion on Datasets}

In general, the clear trend is a growing size of datasets of egocentric videos, and more detailed annotations provided. For example, datasets focused on action recognition and anticipation tend to provide object annotations \citep{damen2018scaling, damen2014you, pirsiavash2012detecting, shan2020understanding}, gaze measurments \citep{tech2018extended, damen2014you, fathi2012learning, jang2019epic, zhang2018coarse} and hand annotations \citep{pirsiavash2012detecting, bambach2015lending, shan2020understanding}. Datasets including untrimmed videos and unscripted activities performed by many subjects in different scenarios, such as the EPIC-KITCHENS dataset series \citep{damen2018scaling, damen2020rescaling} are recently gaining popularity. These datasets are particularly useful to study future prediction tasks due to their ability to capture complex real world scenarios in an unscripted manner.

Yet, some modalities and tasks are to be explored deeper: only 3 datasets provide paired first- and third-person videos \citep{sigurdsson2018charades, elfeki2018third, de2009guide}, of which only CMU-MMAC contains untrimmed sequences of non-scripted activities. However, its size is not sufficient for studying action anticipation, and it is a single-environment dataset. Human location coordinates are rarely provided as labels. While coordinate information is provided in datasets for trajectory prediction, these are missing in popular datasets, so additional efforts are required to recover the camera wearer location from video inputs \citep{nagarajan2020ego}.

Trajectory forecasting datasets are widely represented in the context of autonomous vehicles \citep{yao2019egocentric, poibrenski2020m2p3} for the purpose of developing algorithms for self-driving cars. However, egocentric trajectory forecasting is an underexplored task and only one public dataset among the considered ones suits that purpose \citep{singh2016krishnacam}. 

While the vast majority of datasets consists of videos recorded on purpose, only few contain videos scraped from the Internet \citep{shan2020understanding}. This approach allows to significantly enlarge the size of the datasets, albeit often at the cost of more noise in the data. Another opportunity to increase the size of datasets of egocentric videos comes from setting up simulated 3D environments, as it was implemented in \citep{elfeki2018third,orlando2020egocentric}. 

\section{Models}
\label{sec:models}
This section discusses the main approaches proposed in the literature to anticipate the future from egocentric videos. Specifically, Section~\ref{sec:characteristics} introduces the main characteristics of future prediction methods considered in our analysis, Section~\ref{modalities} further analyses the main input modalities (e.g., RGB frames, optical flow, etc.) considered by the methods, whereas Section~\ref{discussionmodels} reports a comparative discussion on the revised methods, while Section~\ref{sec:evaluation} discusses the approaches used to evaluate and compare models.
Table~\ref{tab:models} provides a concise summary of the reviewed approaches. Where possible, we provide pointers to implementations of methods publicly available.

\subsection{Characteristics of methods}
\label{sec:characteristics}
This section summarises the main attributes according to which the reviewed methods have been categorised, which correspond to columns in Table~\ref{tab:models}. Some of these attributes reflect the parameters introduced in Section~\ref{sec:introduction}, while others, such as the datasets used for testing and the employed modalities, are more technical.

\textit{Task.} The challenges which can be addressed by a model introduced in a paper. Some models have been designed to tackle several tasks, such as action anticipation and predicting anticipation hotspots.
    
\textit{Datasets.} The datasets a given model has been tested on.
It is important to consider this characteristic to assess in which scenarios the method is likely to work best.
    
\textit{Modalities.} A list of modalities used by the future prediction method. Commonly used modalities are discussed in detail in Section \ref{modalities}.
    
\textit{Components.} List of components (e.g., 2D CNNs, recurrent networks, attention mechanisms) included in the model. This attribute indicates the current tendencies in research and highlights which components are most used in methods for
future prediction from egocentric video. A discussion of methods with respect to the used components is reported in \ref{discussionmodels}.
    
\textit{Specific loss.} This parameter indicates whether a method utilise a task-specific loss function for training the models, or loss functions commonly used for generic deep learning tasks (e.g., classification). Specific loss functions are sometimes considered to leverage specific properties of future prediction (e.g., to account for uncertainty) or if auxiliary tasks are included.
    
\textit{End-to-end vs feature-based training.} This attribute indicates whether model was trained end-to-end, or it relied on pre-extracted features. Learning from features allows to save computational resources during training and avoid overfitting.
Nevertheless, the ability to learn or fine-tune a task-specific representation end-to-end may be beneficial depending on the specific task.
Moreover, pre-extracted features are not available in a real-life streaming scenario, so models trained from features should also consider the impact of feature extraction at inference time.
    
\textit{Time-scale of prediction.} 
Whether the method considers the time-scale of the prediction. We suggest that predictive methods consider the time-scale of prediction if multiple subsequent predictions are made (i.g., trajectory forecasting tasks), or if a future event is predicted along with the expected time when the event will take place, thus explicitly answering the ``when?'' question.
    
\textit{Own Actions vs. External events.} Whether the method predicts future activities performed by a camera wearer (own), or by other persons in the scene (external).
    
\textit{Internal state prediction.} Whether the method explicitly predicts its internal state (e.g., a future representation) or makes predictions directly based on a representation of the past.
    

Note that we do not report columns in Table~\ref{tab:models} for the \textit{generalization capabilities} and \textit{focusing capabilities} of predictive systems discussed in Section~\ref{sec:introduction}.
However, \textit{focus capabilities} of a method are related to the presence of attention mechanisms which, if present, are listed in the ``components'' column of Table~\ref{tab:models}.
As for \textit{generalization capabilities}, state-of-the-art approaches for future predictions from egocentric videos have so far considered scenarios in which methods are trained and tested on the same dataset, which is mainly due to the different taxonomies often included in datasets for tasks such as action anticipation. In lack of such systematic analysis, which we believe future works should consider, it is reasonable to assume that methods which are successful when trained and tested on more datasets, as it is reported in the ``datasets'' column of Table~\ref{tab:models}, tend to have better generalisation abilities.

\begin{table*}[!htp]\centering
\caption{Methods for future predictions from egocentric videos. The following abbreviations are used in the table. Task:  Short-Term Action Anticipation (AA), Long-term Action Anticipation (LTAA), Action Recognition (AR), Future Action Captioning (AC), Next Active Object Prediction (NAO), Future Gaze Prediction (GP), Trajectory Forecasting (TF), Action Segmentation (AS), Interaction Hotspots Prediction (IH). Modalities: Optical Flow (OF), Objects (Obj), Bounding Boxes (BB). Components: Recurrent Neural Network (RNN), 2D/3D Convolutional Neural Network (2DCNN/3D CNN), Attention (Att), Modality Attention (Att-mod), Heuristics (Heu), Generative Adversarial Network (GAN), Semantics Modelling (Sem). End-to-end vs feature-based: End-to-end (e2e), feature-based (fea).}
\adjustbox{max width=\textwidth}{
\setlength{\tabcolsep}{2pt}
\scriptsize
\begin{tabular}{|C{2.2cm}|C{1cm}|C{1cm}|C{2.2cm}|C{2.2cm}|C{2.2cm}|C{0.9cm}|C{1cm}|C{1.2cm}|C{1cm}|C{0.9cm}|C{0.9cm}|}\hline
\textbf{Method} &\textbf{Code}  &\textbf{Task} &\textbf{Datasets} &\textbf{Modalities} &\textbf{Components} &\textbf{Specific Loss} &\textbf{End-to-end vs feature-based}&\textbf{Time-Scale} &\textbf{Own/ External} &\textbf{Internal State Prediction}\\ \hline


\cite{soo2016egocentric} &N/A  &TF &EgoMotion &RGB &Heu, 2DCNN &yes &e2e &yes &own &no\\ \hline

\cite{vondrick2016anticipating} &N/A &NAO, AA &ADL &RGB &2DCNN, Heu &no &e2e &no &own/ external &yes\\ \hline

\cite{polatsek2016novelty} &N/A &GP &own dataset &RGB, OF, Gaze &Proba, Att, Heu &no &e2e &no &own &no\\ \hline

\cite{singh2016krishnacam} &N/A &TF &Krishnacam &RGB &Heu, 2DCNN &no &e2e &no &own &no\\ \hline

\cite{furnari2017next} &N/A &NAO &ADL &Obj-BB &Heu, 2DCNN &no &fea &no &own &no\\ \hline

\cite{zhang2017deep} &\href{https://github.com/Mengmi/deepfuturegaze_gan}{Code} &GP & GTEA, GTEA+ &RGB, Gaze &3DCNN, GAN, Att-spatial &no &e2e &yes &own &yes\\ \hline

\cite{wu2017anticipating} &\href{https://github.com/gina9726/Intent-Anticipate}{Code} &AA &Daily Intention Dataset &RGB, Obj, Hand-sensor &RNN, 2DCNN &no &e2e &no &own &no\\ \hline

\cite{su2017predicting} &N/A &TF &own (Basketball Dataset) &RGB, Obj-Bbox, Gaze &RNN, Graphs, Att, Heu, 2DCNN &yes &fea &yes &own/ external &no\\ \hline

\cite{furnari2018leveraging} &\href{https://github.com/fpv-iplab/action-anticipation-losses}{Code} &AA &EPIC-Kitchens &N/A &N/A &yes &N/A &no &own/ external &no\\ \hline

\cite{shen2018egocentric} &N/A &AA, AR & GTEA Gaze, GTEA Gaze + &RGB, Hand-Mask, Gaze &RNN, Att, 2DCNN &no &e2e &no &own &no\\ \hline

\cite{yagi2018future} &\href{https://github.com/takumayagi/fpl}{Code} &TF &First-Person Locomotion (FPL) &RGB, Pose &2DCNN &no &fea &yes &ext &no \\ \hline

\cite{zhang2018anticipating} &\href{https://github.com/Mengmi/deepfuturegaze_gan}{Code} &GP & GTEA, GTEA+, OST, Hollywood2 &RGB, OF &3DCNN, GAN, Att-visual &no &e2e &yes &own &yes\\ \hline

\cite{huang2018predicting} &\href{https://github.com/hyf015/egocentric-gaze-prediction}{Code} &GP & GTEA Gaze, GTEA Gaze + &RGB, OF &3DCNN, RNN, Att &no &fea &no &own &yes \\ \hline

\cite{bertasius2018egocentric} &N/A &TF &Egocentric One-on-One Basketball Dataset &RGB &Heu, 2DCNN &no &N/A &no &own &no \\ \hline

\cite{garcia2018predicting} &\href{https://github.com/GarciaDelMolino/contextual-event-segmentation}{Code} &AS &EDUB-SegDesc &RGB &RNN, 2DCNN &no &fea &no &own &no \\ \hline

\cite{ohn2018personalized} &N/A &TF & own & RGB & RNN, 2DCNN & no & fea &yes &own &no \\ \hline

\cite{fan2018forecasting} &N/A &TF & ADL, Human interaction videos & RGB, OF & 3DCNN & no & e2e &no &own &yes \\ \hline

\cite{liu2019forecasting} &N/A &AA, IH &EGTEA Gaze+, EPIC-Kitchens &RGB &3DCNN, Proba, Att-Spatial &yes &e2e &no &own &yes \\ \hline

\cite{ke2019time} &N/A &AA, LTAA &EPIC-Kitchens, 50Salads &RGB &3DCNN, Att-temporal &yes &fea &yes &own &no\\ \hline

\cite{nagarajan2020ego} &\href{http://vision.cs.utexas.edu/projects/ego-topo/}{Code} &ATAA, AA, AR &EPIC-KITCHENS, EGTEA+ &RGB &3DCNN, Graphs, Heu &no &e2e &no &own &no\\ \hline

\cite{huang2019mutual} &N/A &GP &EGTEA, GTEA Gaze + &RGB, OF &3DCNN, Att &no &e2e &no &own &no\\ \hline

\cite{guan2019generative} &N/A &TF, AA &EPIC-Kitchens &RGB &Proba, Heu, 2DCNN &no &e2e &no &own &no\\ \hline

\cite{miech2019leveraging} &N/A &AA &EPIC-KITCHENS, Breakfast, ActivityNet 200 &RGB &2DCNN &no &fea &no &own &no\\ \hline

\cite{rotondo2019action} &N/A &AA &Stanford-ECM &RGB, Acceleration (OF), Heart rate (hand) &2DCNN, Heu &yes &fea &no &own &no\\ \hline

\cite{nagarajan2019grounded} &\href{https://github.com/Tushar-N/interaction-hotspots}{Code} &IH &OPRA, EPIC-Kitchens &RGB &RNN, Att-visual, 2DCNN &no &e2e &no &own &yes \\ \hline

\cite{chen2019behavioral} & N/A &TF & Stanford 2D-3D-S & RGB, Depth, Location & Graphs, 3DCNN & no & fea &no &own &no\\ \hline

\cite{sener2019zero} &N/A &AC &own (Tasty Videos) &RGB &RNN, Sem, 2DCNN &no &fea &no &own &no\\ \hline

\cite{tavakoli2019digging} &N/A &GP &USC, GTEA Gaze, GTEA Gaze + &RGB, OF, Hand &RNN, Att-spatial, 2DCNN &no &fea &no &own &no\\ \hline

\cite{furnari2019would} &\href{https://github.com/fpv-iplab/rulstm}{Code} &AA, AR &EPIC-KITCHENS, EGTEA Gaze + &RGB, OF, Obj &RNN, Att-Mod, 2DCNN &no &fea &no &own &no\\ \hline

\cite{camporese2020knowledge} &N/A &AA, AR &EPIC-KITCHENS &RGB, OF, Obj &RNN, Att-Mod, Sem, 2DCNN &yes &fea &no &own &no\\ \hline

\cite{sener2020temporal} &N/A &AA &Breakfast Actions, 50Salads, EPIC-Kitchens &RGB, OF, Obj &3DCNN, RNN, Att &no &fea &no &own &no \\ \hline

\cite{zhang2020egocentric} &N/A &AA &EPIC-Kitchens &RGB, OF, Obj &RNN, Att-mod, Sem, 2DCNN &no &fea &no &own &no \\ \hline

\cite{zhaodiverse} &N/A &AA &EPIC-Kitchens, Breakfast, 50Salads &RGB &GAN, RNN, Sem, 2DCNN &yes &e2e &yes &own &yes\\ \hline

\cite{aakur2020unsupervised} &\href{https://saakur.github.io/Projects/GazePrediction/#}{Code} &GP & GTEA, GTEA Gaze + &RGB, OF &Proba, Heu, 2DCNN &no &e2e &no &own &no\\ \hline

\cite{dessalene2020egocentric} &N/A &AA & EPIC-KITCHENS &RGB, Obj &LSTM, Graph, 3DCNN, Sem &no &fea &no &own &yes\\ \hline




\end{tabular}
}
\label{tab:models}
\end{table*}

\subsection{Modalities}\label{modalities}

Due to the dynamic nature of egocentric videos and to the indirect relationships between past observations and future events, methods usually benefit from additional modalities other than RGB alone \citep{furnari2019would}. These can be mid-level representations of RGB frames themselves (e.g., optical flow), high-level representations (e.g., hands, object bounding boxes), or signals collected with additional sensors (e.g., depth, gaze, camera coordinates). In the following, we revise the main modalities which are beneficial when making predictions about the future from egocentric video.



\textbf{Optical flow} \citep{horn1981determining} represents the change in the appearance of frames by reporting the displacement of pixel intensity values over time. The motion features captured by optical flow provide dynamic information, and so optical flow is widely used in various video processing tasks such as motion detection \citep{shafie2009motion, loy2012salient} and action recognition \citep{wang2016temporal, simonyan2014two}. Optical-flow is a frequently used modality in future prediction methods. However, calculating optical flow on-the-fly is computationally demanding, so the majority of models use pre-computed optical flow features for training \citep{furnari2019would, huang2018predicting, tavakoli2019digging, camporese2020knowledge}. However, it should be noted that the computation of optical flow still needs to be computed at inference time, which limits its use in real-time applications. Different algorithms can be used to compute optical flow. The most common approaches are TVL1 \citep{zach2007duality}, FlowNet 2.0 \citep{ilg2017flownet}, LDOF \citep{brox2010large}, the Horn-Schunck algorithm \citep{horn1981determining}, and SMD (Salient Motion Detection) \citep{loy2012salient}. The adoption of this wide range of methods indicates a lack of consensus among researchers on which algorithm to use for calculating optical flow. The oldest method still used in modern papers is dated 1981 (Horn-Schunck algorithm), whereas new solutions are also starting to appear, e.g., FlowNet 2.0 \citep{ilg2017flownet}.

\textbf{Hands}. Hands representations play an important role in action recognition \citep{kapidis2019egocentric, zhou2016cascaded, tekin2019h+, ma2016going, singh2016first, kapidis2019multitask}. However they are not as widely used in anticipation \citep{shen2018egocentric, tavakoli2019digging}. This may suggest that leveraging such representations is not straightforward for predicting future actions. This may be due to the fact that hand position changes are likely to be observed just before the beginning of the action, which makes them hard to be used effectively for mid- and long-term future prediction. Nevertheless, it should be noted that a thorough investigation on the role of hands in future action prediction is still missing. Although most models used visual representations of hands poses, such as bounding boxes or masks, sensor data may also be used. For instance, heart rate is measured by a hand-mounted sensor in \citep{rotondo2019action}, while in \citep{wu2017anticipating} a hand-mounted accelerometer and a camera are exploited to anticipate user's intentions. An extensive survey on using hands for modelling first-person vision scenarios is provided in \citep{bandini2020analysis}, although this survey does not focus on challenges related to future predictions.

\textbf{Objects}. Items surrounding the camera wearer give a strong clue about the actor's possible future actions and intentions. Hence, extracting object representation from original RGB frames may be useful to indicate the next object of interaction. Several papers used simple bag-of-words representations of objects present in the frame \citep{furnari2019would, camporese2020knowledge, sener2020temporal}, while others encoded objects in bounding-box representations \citep{su2017predicting}, or even just in coordinates of bounding box centres \citep{furnari2017next}. Detecting objects in real time is computationally expensive, so many approaches utilise pre-trained object detectors at training time, such as Faster RCNN \citep{ren2015faster} with ResNet-101 backbone \citep{he2016deep} in \citep{furnari2019would, camporese2020knowledge}. Similarly to optical flow, however, the overhead to detect objects should also be considered at inference time.

\textbf{Gaze}. Human gaze can serve as a strong clue on what a user points their attention to, and so it may be a strong indicator of what they will do next. Using gaze data as input requires wearing a gaze tracker. These devices are not of high accessibility for a wide range of users, and only few datasets provide gaze annotation. In \citep{shen2018egocentric}, gaze serves as an attention weighting mechanism for predicting the next actions. The authors of \citep{huang2018predicting, tavakoli2019digging} used gaze fixations as supporting input to RGB and optical flow data and shown that jointly learning the 3 modalities allows to predict the next fixation point. As shown in \citep{zhang2018coarse}, egocentric gaze prediction can be of particular interest in psychology research, especially for behavioural modelling.

\subsection{Discussion on Models}\label{discussionmodels}

\begin{table*}[t]
	\caption{Egocentric action anticipation results on the EPIC-KITCHENS-55 test set.}
	\label{tab:anticipation_ek_test}
	\adjustbox{max width=\textwidth}{
		\setlength{\tabcolsep}{3pt}
		\begin{tabular}{llccc|ccc|ccc|ccc}
			& & \multicolumn{3}{c|}{Top-1 Accuracy\%} & \multicolumn{3}{c|}{Top-5 Accuracy\%} & \multicolumn{3}{c|}{Avg Class Precision\%} & \multicolumn{3}{c}{Avg Class Recall\%} \\ \hline
			& & VERB & NOUN & ACTION & VERB & NOUN & ACTION & VERB & NOUN & ACTION & VERB & NOUN & ACTION \\ \hline
			\multirow{12}{*}{\rotatebox{90}{\textbf{S1}}} &
			{DMR~\cite{vondrick2016anticipating}} & 26.53 & 10.43 & 01.27 & 73.30 & 28.86 & 07.17 & 06.13 & 04.67 & 00.33 & 05.22 & 05.59 & 00.47\\
			&{ED~\cite{gao2017red}} & 29.35 & 16.07 & 08.08 & 74.49 & 38.83 & 18.19 & 18.08 & 16.37 & 05.69 & 13.58 & 14.62 & 04.33\\
			&
			{2SCNN~\cite{damen2018scaling}} & 29.76 & 15.15 & 04.32 & 76.03 & 38.56 & 15.21 & 13.76 & 17.19 & 02.48 & 07.32 & 10.72 & 01.81\\
			&ATSN~\cite{damen2018scaling} & 31.81 & 16.22 & 06.00 & 76.56 & 42.15 & 28.21 & 23.91 & 19.13 & 03.13 & 09.33 & 11.93 & 02.39\\
			&MCE~\cite{furnari2018leveraging} & 27.92 & 16.09 & 10.76 & 73.59 & 39.32 & 25.28 & 23.43 & 17.53 & 06.05 & 14.79 & 11.65 & 05.11\\
			
			&{\cite{miech2019leveraging}} & 30.74 & 16.47 & 09.74 & 76.21 & 42.72 & 25.44 & 12.42 & 16.67 & 03.67 & 08.80 & 12.66 & 03.85\\
			&\cite{furnari2019would} & 
			33.04 & 22.78 & 14.39 & 79.55 & 50.95 & 33.73 & 25.50 & 24.12 & 07.37 & 15.73 & 19.81 & 07.66\\
			
			&\cite{camporese2020knowledge} &35.04 &23.03 &14.43 &79.56 &52.90 &34.99 &30.65 &24.29 &06.64 &14.85 &20.91 &07.61\\
			
			&\cite{liu2020forecasting} &36.25 &23.83 &15.42 & 79.15 &51.98 &34.29 &24.90 &24.03 &06.93 &15.31 &21.91 & 07.88\\ 
			&\cite{wu2020learning} & 35.44 & 22.79 & 14.66 & 79.72 & 52.09 & 34.98 & 28.04 & 24.18 & 06.66 & \textbf{16.03} & 19.61 & 07.08\\ 
			&\cite{dessalene2020egocentric} & 32.20 & \textbf{24.90} & 16.02 & 77.42 & 50.24 & 34.53 & 14.92 & 23.25 & 04.03 & 15.48 & 19.16 & 05.36\\ 
			&\cite{sener2020temporal} & \textbf{37.87} & 24.10 & \textbf{16.64} & \textbf{79.74} & \textbf{53.98}  & \textbf{36.06} & \textbf{36.41} & \textbf{25.20} & \textbf{09.64} & 15.67 & \textbf{22.01} & \textbf{10.05}\\
			
			\hline

			\hline
			\multirow{12}{*}{\rotatebox{90}{\textbf{S2}}} &
			{DMR~\cite{vondrick2016anticipating}} & 24.79 & 08.12 & 00.55 & 64.76 & 20.19 & 04.39 & 09.18 & 01.15 & 00.55 & 05.39 & 04.03 & 00.20\\
			&{ED~\cite{gao2017red}} & 22.52 & 07.81 & 02.65 & 62.65 & 21.42 & 07.57 & 07.91 & 05.77 & 01.35 & 06.67 & 05.63 & 01.38\\
			&{2SCNN~\cite{damen2018scaling}} & 25.23 & 09.97 & 02.29 & {68.66} & 27.38 & 09.35 & 16.37 & 06.98 & 00.85 & 05.80 & 06.37 & 01.14\\
			&ATSN~\cite{damen2018scaling} & {25.30} & {10.41} & 02.39 & 68.32 & {29.50} & 06.63 & 07.63 & 08.79 & 00.80 & 06.06 & {06.74} & 01.07\\
			&MCE~\cite{furnari2018leveraging} & 21.27 & 09.90 & {05.57} & 63.33 & 25.50 & {15.71} & 10.02 & 06.88 & {01.99} & {07.68} & 06.61 & {02.39}\\
			
			&{Miech et al.\cite{miech2019leveraging}} & 28.37 & 12.43 & 07.24 & 69.96 & 32.20 & 19.29 & 11.62 & 08.36 &02.20 & 07.80 & 09.94 & 03.36\\

			&\cite{furnari2019would} & 27.01 & 15.19 & 08.16 & 69.55 & 34.38 & 21.10 & {13.69} & 09.87 & 03.64 & 09.21 & 11.97 & 04.83\\
			
			&\cite{camporese2020knowledge} &29.29  &15.33  &08.81  &70.71  &36.63  &21.34  &14.66  &09.86  &04.48  &08.95  &12.36  &04.78\\
			
			& \cite{liu2020forecasting} & \textbf{29.87} & 16.80 & 9.94 & \textbf{71.77} & \textbf{38.96} &	23.69 &	15.96 &	12.02 & 04.40 & 09.65 & 13.51 & 05.18\\ 
			&\cite{wu2020learning} & 29.33 & 15.50 & 9.25 & 70.67 & 35.78 & 22.19 & 17.10 & 12.20 & 03.47 & 09.66 & 12.36 & 05.21\\
			&\cite{dessalene2020egocentric} & 27.42 & \textbf{17.65} & \textbf{11.81} & 68.59 & 37.93 & \textbf{23.76} & 13.36 & \textbf{15.19} & 04.52 & \textbf{10.99} & \textbf{14.34} & 05.65 \\
			&\cite{sener2020temporal} & 29.50 & 16.52 & 10.04 & 70.13 & 37.83 & 23.42 & \textbf{20.43} & 12.95 & \textbf{04.92} & 08.03 & 12.84 & \textbf{06.26}\\
			
			\hline
			
			\hline
			
		\end{tabular}
		}
\end{table*}




Most methods make use of a pre-trained 2DCNN backbone as a feature extractor or as an object detector, as shown in Table~\ref{tab:models}.
Features extracted by 2D CNNs models can be aggregated using attention mechanisms~\citep{wu2019long,sener2020temporal} or recurrent neural networks (RNNs) such as LSTMs \citep{hochreiter1997long} or GRUs \citep{chung2014empirical}, as done in~\citep{shen2018egocentric, furnari2019would, wu2017anticipating, tavakoli2019digging}.

Spatio-temporal (3D) convolutions are also used for future predictions~\citep{liu2019forecasting} and play an important role in models based on generative adversarial networks (GAN) \citep{goodfellow2014generative}. 
For instance, \citep{elfeki2018third} used 3DCNN GANs to generate first-person video from third-person camera view, and in \citep{zhang2017deep, zhang2018anticipating} spatio-temporal representations are used to generate the future gaze heatmaps. The combination of 2 convolutional networks, one for predicting the future trajectory and another for defining camera position and orientation, has been proposed in \citep{bertasius2018egocentric}.

Although the deep learning paradigm is dominant in computer vision methods, shallow modules and heuristics still find their use in the egocentric future anticipation approaches. The authors of \citep{furnari2017next} trained a random forest \citep{ho1998random} to classify active vs passive object trajectories and predict whether or not an object will be involved in a future action. 
The authors of \citep{soo2016egocentric} built on EgoRetinal map \citep{choset2005principles} and performed geometrical modelling to obtain task-specific features for trajectory prediction. 
The model presented in \citep{singh2016krishnacam} estimates the wearer’s future trajectory as the average of the trajectories of its top-10 nearest neighbours. 

Predicting the future from egocentric videos may also be considered from the perspective of sequence modelling. In that context, methods originated from classic sequence modelling tasks such as time-series analysis or natural language processing can be exploited. Recurrent neural networks have been mainstream in the language modelling landscape for several years \citep{goldberg2017neural}, and they have been adapted to video processing~\citep{sun2019videobert}. 
In several papers, recurrent neural networks are used in a straightforward way: a single RNN serves as a video aggregation module \citep{nagarajan2019grounded, wu2017anticipating, sener2019zero}. The authors of \citep{garcia2018predicting} utilize 2-layer LSTMs to perform action segmentation on video stream. RU-LSTM \citep{furnari2019would} consists of a cascade of 2 LSTMs: a rolling LSTM for aggregating the past features and an unrolling LSTM to predict future actions. The authors of \citep{shen2018egocentric} consider 2 LSTM-based modules: 1) an asynchronous module which reasons about the temporal dependency between events and 2) a synchronous module which softly attends to informative temporal durations for more compact and discriminative feature extraction. In \citep{huang2018predicting}, a LSTM is used for modelling gaze fixation points. Although not tested on egocentric video datasets, the model introduced in \citep{farha2020long} solves the action anticipation and segmentation tasks by introducing a cascade of GRU networks and cycle consistency loss which ensures that predictions are matched semantically with observable frames.

Recurrent neural networks, and in particular LSTM networks, have certain limitations, which restricted them from remaining the dominating models in the sequence processing landscape. RNNs are difficult to train mostly due to the very long gradient paths. Transfer learning is limited with recurrent neural networks, so specific and large labelled datasets are required for every new task. With the introduction of transformers \citep{vaswani2017attention}, these limitations were overcome at least partially. Transformer-based architectures \citep{devlin2018bert, peters2018deep, yang2019xlnet} showed good results in various language modelling tasks, such as language translation, question answering and text classification. Transformers have also been adapted to the video understanding scenario. VideoBERT \citep{sun2019videobert} was designed to text-to-video generation and future forecasting. Action transformer models for recognising and localizing human actions in video clips from third-person view have been introduced in  \citep{girdhar2019video}. Transformer-based models are efficient not only in recognising activities in given video segments, but also in predicting the future. The TTPP model \citep{wang2020ttpp} serves for action anticipation and Trajectory-Transformer \citep{giuliari2020transformer} serves for predicting where a person on a video will go. Although most of these results have been reported on third person vision datasets, it is likely that transformer networks can be exploited in the egocentric scenario.

Learning to attend to a specific region, timestep or modality is important for recognition and future prediction. In \citep{zhang2017deep, zhang2018anticipating, tavakoli2019digging}, spatial attention has been used to predict the future gaze, while in \citep{liu2019forecasting, nagarajan2019grounded}, it has been used as a mechanism to solve the task of predicting interaction hotspots. Spatial attention mechanisms can be of 2 types: stochastic (hard) and deterministic (soft) attention. These two approaches differ in the way in which attention maps are generated \citep{xu2015show} and are both possible in the video processing scenario. Temporal attention helps in processing long sequences by attending to a frame or feature located at a specifc temporal location. The authors of \citep{ke2019time} proposed a time-conditioned approach for long-term action anticipation which included an attended temporal feature and a time-conditioned skip connection to extract relevant information from video observations. However, time and space are not the only dimensions with respect to which attention can be used. Indeed, when working with multimodal data, classic early- and late-fusion approaches may be sub-optimal as shown in \citep{mees2016choosing}, so it may be beneficial to fuse features from different modalities by weighting them using modality-specific weight, as implemented, for example in \citep{furnari2019would} for merging RGB, optical flow and object-based predictions. The model introduced in \citep{zhang2020egocentric} fuses an ``intuition-based prediction network'', imitating implicit intuitive thinking process, and an ``analysis-based prediction network'', which processes information under designed rules with a similar modality attention mechanism.

\subsection{Evaluation protocols and performance comparison} \label{sec:evaluation}

{Evaluation protocols for future predictions should take into account the intrinsic uncertainty of future events. Indeed, evaluation measures for future prediction models often follow a ``best of many'' criterion in which the model is allowed to make a plausible predictions for a given input example. The best prediction (i.e., the one giving the best performance) is then selected for the evaluation. For example, \citep{koppula2015anticipating} evaluates an example as a true positive, if the correct class is among the top 3 predictions, \citep{park2016egocentric} consider the predicted trajectory giving the smallest error among the top 10 trajectories, \citep{damen2018scaling} evaluate action anticipation using top-5 accuracy and \citep{furnari2018leveraging} introduced mean top-5 recall.}

{Egocentric trajectory forecasting papers rely on \textit{end point displacement error} and \textit{average displacement error} to evaluate predicted trajectory against the ground truth \citep{ohn2018personalized}. Papers focused on hands, objects and other persons' trajectory forecasting may utilise metrics for object detection, such as intersection over union (IoU) or mean average precision (mAP) \citep{fan2018forecasting}. It is worth noting that only egocentric action anticipation is a well established task, with a dedicated benchmark (EPIC-KITCHENS), and a challenge organised each year \citep{damen2018scaling, damen2020rescaling}. This contributed to the emergence of shared evaluation protocols which enable comparisons between different works.}

{To give a snapshot of the current status of the research in this field, Table~5 reports publicly available results of action anticipation methods on EPIC-KITCHENS-55. The results indicate that models designed for third-person action recognition, such as 2SCNN~\citep{damen2018scaling} or ATSN~\citep{damen2018scaling} can only serve as a baseline for egocentric activity anticipation, achieving limited performance. Approaches designed specifically for egocentric video processing and for future prediction, such as \citep{furnari2019would, wu2020learning, sener2020temporal, dessalene2020egocentric} achieve better results. Another noticeable feature arising from the analysis of the Table~5 is a difference in models performance on the set of seen (S1) and unseen (S2) kitchens (indicating whether the test environments were present in the training set), showing that models tend to overfit to seen environments, and thus more testing should be performed before applying models to real-world setting. Overall, the Table~~\ref{tab:anticipation_ek_test} shows steady improvement of results over the time, which can indicate the importance of having an established benchmark for the challenge.}

{It is also worth noticing that current trends in deep learning, such as using attention mechanisms instead of recurrent models for sequence modelling \citep{vaswani2017attention}, or graph convolutional networks \citep{zhang2019graph}, gave inspiration to state-of-the-art models \citep{sener2020temporal, dessalene2020egocentric}. Thus, although architectures designed for third-person action recognition do not achieve high scores on activity anticipation task, some ideas and model components could be successfully adapted from related research in machine learning.}

\section{Conclusion and Further Research}

In this survey, we have summarized the current state of research in future prediction from egocentric vision. Looking into the future is necessary when it comes to safety and assistance. In particular, egocentric future prediction algorithms may recognize dangerous patterns of executed actions in industrial production workflows or in everyday routine and send a signal to prevent a camera wearer from performing unsafe activities. Supporting people with cognitive problems or blind people by understanding their intentions and providing assistance without an explicit request is also a task which can benefit from egocentric future prediction algorithms. Finally, tasks as egocentric trajectory forecasting or future gaze prediction are relevant for marketing purposes, for example, when modelling customers' behaviours and navigating them through a shopping center. Moreover, the devices which allow to collect egocentric visual data, process it and provide assistance to the user are already available on the market. Smart glasses with augmented reality technology can be used for navigating an environment, highlighting specific objects with AR, or providing voice assistance. 

{However, currently there is a lack of unified benchmarks for building, evaluating and comparing the methods aimed to future prediction. Among the challenges discussed in Section~\ref{sec:challenges}, only short-term action anticipation has an associated benchmark~\citep{damen2018scaling}, while the only challenge focusing on trajectory forecasting is currently related to the TrajNet~\citep{sadeghian2018trajnet}, which focuses on the third-person scenario. As discussed in Section~\ref{sec:models}, the introduction of the Epic Kitchens Action Anticipation Challenge in 2018 and the associated competition boosted research in egocentric future prediction. Similarly, we believe that establishing unified benchmarks also for other challenges of egocentric future prediction, such as egocentric trajectory forecasting or long-term action anticipation, could raise the awareness and focus of the academia and industry in these fields.}

{Besides, current benchmarks (e.g., EPIC-KITCHENS) focus on a very narrow scenario in a single environment type (e.g., kitchens) and do not capture a complete snapshot of a person’s life. Hence, there is no guarantee that a method trained on such a benchmark could generalize to other environments and settings. Thus we believe, that more efforts should be devoted to scaling the environments and verifying applicability of future prediction models in different settings.}

{Another limitation of current benchmarks and datasets is a single-task orientation. Indeed, current benchmarks tend to focus on a single task at a time (e.g., EPIC-KITCHENS is providing annotations for action anticipation, but not for trajectory forecasting or for next active object prediction). It would be desirable to have a large-scale benchmark considering the different challenges so that one task can benefit from the other and we the research community could have a common way to measure the advancement of future predictions in egocentric vision.}


Besides that, currently available datasets of egocentric videos are mostly focusing on everyday activities, such as cooking, walking, etc., and more specific activities that can be of particular interest for industrial applications or personal health assistance are under-explored.

There is also a lack of paired first- and third-person vision datasets, which may be useful for two reasons. The first one is pretty straightforward: many real-life applications may exploit two views from static and wearable cameras. For instance, in assistive robotics, one camera may be fixed or attached to a robot or autonomous assistant to collect third-person vision data, while an egocentric camera worn by a human may stream visual data to the robot which may use this data to predict the human's intentions. Third-person video may also help in identifying the human location, pose, as well as their movements. Thus a combination of a third-person and egocentric view could provide better understanding of the activities currently performed by a human and by that means improve the quality of future predictions. The second motivation for having large-scale paired datasets is related to the possibility to learn a correspondence between views and have a better understanding of a scene even if only one view is provided.

Another possible direction for dataset collection is acquiring egocentric data where several persons are present in the video or even several persons record videos which may also be temporally aligned. Currently, the biggest egocentric datasets focus on activities performed by one person at a time and do not model human interactions.

Depth is an under-explored modality in egocentric future prediction research. While in third-person vision, there are numerous datasets collected with RGB-D cameras (e.g., datasets for autonomous driving cars contain either depth or lidar point clouds), most of the egocentric datasets collected using wearable devices do not include depth data. Information on the distance to the objects can be useful for estimating the time-to-interaction with an object.

From the algorithmic point of view, we believe that more attention should be payed to the inference time of models. Algorithms designed for future prediction should be fast enough to predict future events before they occur, so that they could be applicable in real-time streaming scenarios. Thus, reporting inference time should become a standard practice for works focused on future prediction tasks.

Finally, one possibility for the development of methods designed for egocentric future prediction may lie in the use of unsupervised and semi-supervised learning. From the psychological and physiological research \citep{zador2019critique}, it is clear that learning end-to-end from labels is not the only way the human brain works. So introducing algorithms capable of leveraging also unsupervised or self-supervised observations, may be an interesting direction for further research.

\section{Acknowledgements}

This research has been supported by Marie Skłodowska-Curie Innovative Training Networks - European Industrial Doctorates - PhilHumans Project (\url{http://www.philhumans.eu}), MIUR AIM - Attrazione e Mobilita Internazionale Linea 1 - AIM1893589 - CUP: E64118002540007 and by MISE - PON I\&C 2014-2020 - Progetto ENIGMA - Prog n. F/190050/02/X44 – CUP: B61B19000520008.

\balance
\bibliographystyle{elsarticle-num-names}
\bibliography{refs}

\end{document}